# ArmorCLIP: A Hybrid Defense Strategy for Boosting Adversarial Robustness in Vision-Language Models

Yijun Li[a], Yuhan Liang[b], Yumeng Niu[c], Qianhe Shen[d], Hangyu Liu[e]
[a] Computer Science and Technology, Harbin Institute of Technology, 518055, Shenzhen, China; [b] Faculty of Engineering, University of Bristol, BS8 1TH, Bristol, UK; [c] Computer Science and Technology, Beijing University of Civil Engineering and Architecture, 102612, Beijing, China; [d] Computer Science, Mathematics & Statistics, University of Toronto Mississauga, Mississauga, L5L 1C6, Canada; [e]Xi’an Tieyi ICC, 710054, Xian, Shanxi
lic87876@gmail.com, an22874@bristol.ac.uk, 202204030117@stu.bucea.edu.cn, peter.shenqh@outlook.com, liuhangyu2025@163.com

## ABSTRACT

The robustness of Vision-Language Models (VLMs) such as CLIP is critical for their deployment in safety-critical applications like autonomous driving, healthcare diagnostics, and security systems, where accurate interpretation of visual and textual data is essential. However, these models are highly susceptible to adversarial attacks, which can severely compromise their performance and reliability in real-world scenarios. Previous methods have primarily focused on improving robustness through adversarial training and generating adversarial examples using models like FGSM, AutoAttack, and DeepFool. However, these approaches often rely on strong assumptions, such as fixed perturbation norms or predefined attack patterns, and involve high computational complexity, making them challenging to implement in practical settings. In this paper, we propose a novel adversarial training framework that integrates multiple attack strategies and advanced machine learning techniques to significantly enhance the robustness of VLMs against a broad range of adversarial attacks. Experiments conducted on real-world datasets, including CIFAR-10 and CIFAR-100, demonstrate that the proposed method significantly enhances model robustness. The fine-tuned CLIP model achieved an accuracy of 43.5% on adversarially perturbed images, compared to only 4% for the baseline model. The neural network model achieved a high accuracy of 98% in these challenging classification tasks, while the XGBoost model reached a success rate of 85.26% in prediction tasks.

## 1. INTRODUCTION

### 1.1. Importance of the Problem

The deployment of Vision-Language Models (VLMs), such as CLIP, in critical real-world applications like autonomous driving, healthcare diagnostics, and security systems underscores the importance of ensuring their robustness. These models must accurately interpret and associate visual and textual data to function effectively. However, their susceptibility to adversarial attacks presents a significant challenge. Adversarial attacks exploit vulnerabilities in these models, potentially causing them to misinterpret data, which could lead to severe consequences in safety-critical applications [1]. Thus, enhancing the robustness of VLMs against such attacks is crucial for maintaining the reliability and safety of systems that rely on them.

### 1.2. Categories of Existing Methods

To mitigate the adversarial vulnerability of deep learning models, two primary defense strategies have been developed: adversarial training and ensemble learning methods.

**Adversarial Training:** This method trains models on both clean and adversarial examples to enhance robustness. While it has shown some effectiveness, it is computationally expensive and often fails to generalize to novel or adaptive attack strategies, limiting its practical applicability in diverse real-world scenarios [2].

**Ensemble Learning Methods:** Techniques such as boosting (e.g., XGBoost, LightGBM) aim to improve robustness by combining multiple models. Although ensemble methods can increase resilience by aggregating diverse predictions, they also face limitations. High computational costs and vulnerability to adaptive adversarial strategies that target the ensemble mechanism itself are significant drawbacks. Additionally, the effectiveness of ensemble methods depends on the diversity and independence of the models used, which can be challenging to achieve consistently.

Despite these advancements, existing methods often rely on specific assumptions, such as fixed perturbation norms or predefined attack patterns, which may not hold in real-world scenarios where adversaries use unexpected or sophisticated strategies.

### 1.3. Our Contributions

This research introduces a novel approach to improving the adversarial robustness of VLMs by integrating advanced adversarial training methods with innovative machine-learning techniques. We combine multiple attack strategies, including FGSM, AutoAttack, and DeepFool, to develop a hybrid model configuration that enhances model resilience. Additionally, we employ ensemble learning methods such as XGBoost and LightGBM to evaluate and improve the detection and classification of adversarial examples. These methods are specifically focused on identifying instances where models like CLIP initially fail.

The main contributions of our research are summarized as follows:

**Methodological Innovation:** Development of a hybrid adversarial training framework that leverages multiple model configurations to bolster robustness against diverse adversarial attacks.

**Algorithmic Advancement:** Application of XGBoost and LightGBM for enhanced adversarial example detection and classification, dynamically adapting model defenses based on the nature of the attack.

**Empirical Validation:** Extensive testing on multiple datasets demonstrates the improved robustness and accuracy of the enhanced CLIP model against various adversarial attacks.

### 1.4. Thesis Layout

The thesis is structured as follows: Section 2 reviews related work in adversarial attacks and defenses in deep learning, as well as applications of VLMs and robustness challenges. Section 3 details our methodology, including the attack models, adversarial training strategies, and integration of machine learning techniques. Section 4 describes the experimental setup and evaluation protocols. Section 5 presents our findings, comparing the performance of baseline and enhanced models. Finally, Section 6 concludes with a discussion of the implications of our research, and Section 7 indicates directions for future work.

## 2. RELATED WORKS

### 2.1. Adversarial Attacks and Defenses in Deep Learning

Recent advancements in deep learning have highlighted the importance of developing robust models that can resist adversarial attacks. Traditional defenses, such as adversarial training and defensive distillation, have been explored extensively. However, these methods are often limited by high computational costs and the challenge of generalizing to novel attacks. Researchers have continued to explore new strategies, including input transformations and hybrid defense mechanisms, to address these limitations [3].

### 2.2. Applications of VLMs and Robustness Challenges

Recent studies have explored using advanced machine learning techniques, such as ensemble learning and boosting methods, to enhance model robustness against adversarial attacks. For instance, XGBoost and LightGBM have been employed to detect and classify adversarial examples effectively, providing a robust mechanism to safeguard against such attacks [4]. These models, by aggregating the strengths of multiple learners, can potentially mitigate the impact of adversarial perturbations by improving the model's overall robustness.

### 2.3. Machine Learning Techniques for Robustness Enhancement

Ensemble learning techniques, including boosting algorithms like XGBoost and LightGBM, have shown promise in improving model robustness against adversarial attacks. By combining multiple weak learners into a single, stronger model, these techniques can reduce the vulnerability of individual models to adversarial examples [5]. However, the effectiveness of ensemble methods hinges on the diversity and independence of the models used, which can be challenging to maintain consistently. This research builds upon these insights by integrating ensemble learning with adversarial training to enhance VLM robustness.

### 2.4. Three Attack Models

To assess the robustness of VLMs, we utilized three well-known adversarial attack methods:

**Fast Gradient Sign Method (FGSM):** A gradient-based attack that generates adversarial examples by applying a perturbation in the direction of the gradient of the loss function aimed at maximizing the model's prediction error [1].

**DeepFool:** An iterative method that calculates the minimal perturbation needed to move an input image across the model's decision boundary, resulting in misclassification [6].

**AutoAttack:** An automated attack framework that combines multiple strategies to generate robust adversarial examples, providing a comprehensive evaluation of model robustness [7].

By combining these diverse attack methods, we aim to thoroughly evaluate the robustness of VLMs, particularly under adversarial conditions, and enhance their resilience through advanced adversarial training and ensemble learning techniques.

# 3. METHOD

## 3.1. Problem Setup

In this study, we focus on the problem of classifying adversarially perturbed images using various machine learning models. Let X represent the input images, Y the true labels, and $\hat{Y}$ the predicted labels. The dataset consists of N samples, where each sample $x_i \in X$ is an image, and $y_i \in Y$ is its corresponding label. The objective is to minimize the misclassification rate under adversarial conditions, which introduces significant challenges in maintaining the robustness of the classification models.

Additionally, the models are also evaluated on their ability to predict the labels $\tilde{Y}$ for new, particularly under adversarial conditions.This extends the study to assess the generalization capabilities of the models when predicting labels for new inputs that may also be adversarially perturbed. This problem is a standard classification task with added complexity due to adversarial attacks, following the framework outlined in previous studies.

## 3.2. Adversarial Training ---CLIP

In this study, we extended the capabilities of the Contrastive Language-Image Pretraining (CLIP) model by subjecting it to adversarial training aimed at improving its robustness against adversarial examples. The CLIP model, introduced by Radford et al., leverages a large-scale dataset of images and textual descriptions to learn visual concepts in a zero-shot manner, enabling it to perform various vision-language tasks without task-specific fine-tuning [8].

As shown in Figure 1, the samples used for adversarial training were generated by two different adversarial models, sequential attack integration and feature-level attack fusion. The generated samples were then combined into a new dataset for adversarial training of CLIP. The adversarial training process consists of two parts: the first step is to retrain CLIP, and the second step is evaluation.

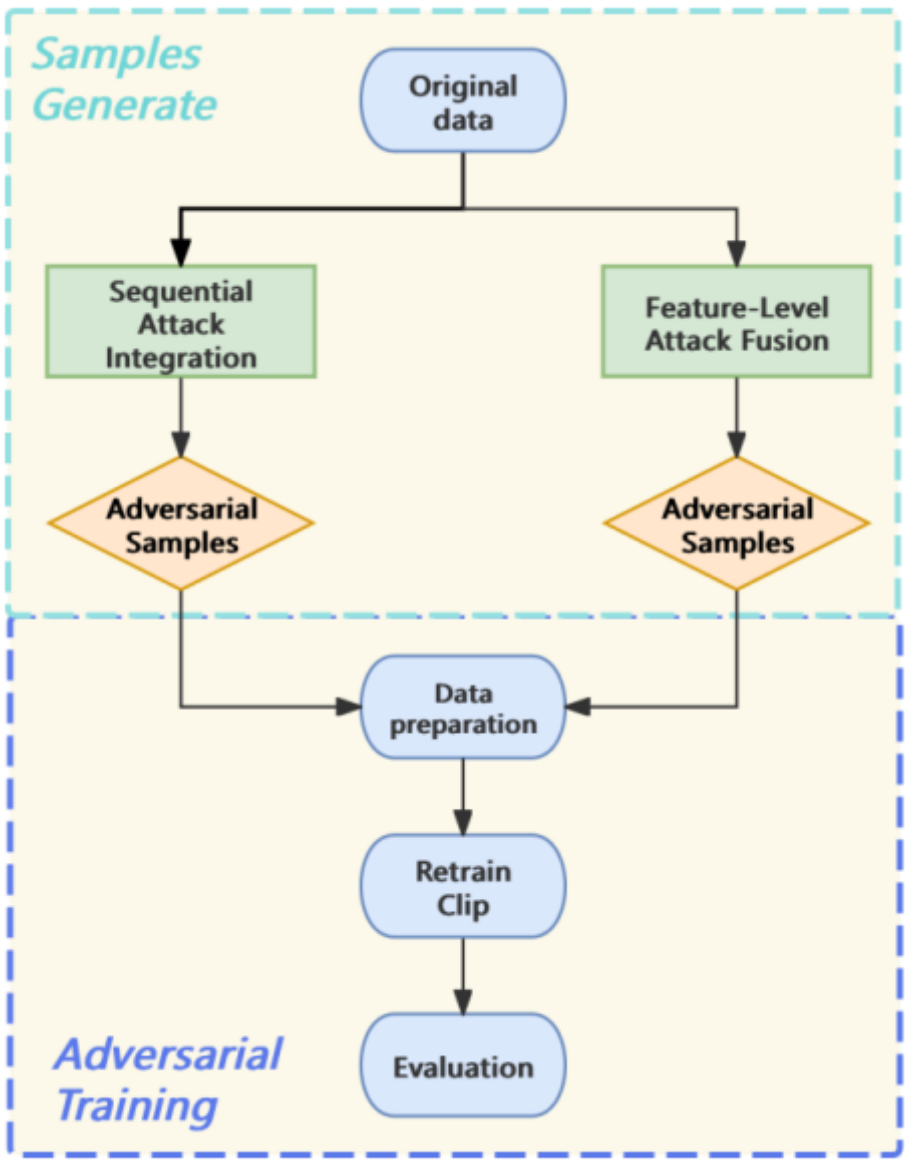

Figure 1 Adversarial training flowchart

### 3.1.1. Sequential Attack Integration

Sequential Attack Integration is a method where multiple attack strategies are applied in a sequence, progressively strengthening the adversarial perturbation. This multi-step approach ensures that the final adversarial example is more effective at circumventing model defenses.

Initial Perturbation: Apply FGSM to generate a base adversarial example $x'$.

Further Perturbation: Refine x′ using DeepFool to reduce the perturbation size while maintaining its adversarial properties.

Final Enhancement: Use AutoAttack to further optimize x″, ensuring robustness against diverse defenses.

### 3.1.2. Feature-Level Attack Fusion

Feature-Level Attack Fusion is an advanced technique that synthesizes adversarial examples by combining features extracted from different attack-generated perturbations. This approach aims to leverage the strengths of each attack to produce a composite adversarial example that is more resilient to model defenses.

**Independent Adversarial Example Generation:** Generate adversarial examples x′FGSM, x′Deepfool, and x′Autoattack using the respective methods.

**Feature Fusion:** the features extracted from different adversarial examples are weighted and combined to generate a composite feature representation:

$$x'_{\text{combined}} = \frac{x'_{\text{FGSM}} + x'_{\text{DeepFool}} + x'_{\text{AutoAttack}}}{3}$$

### 3.1.3. Model Retraining

The model retraining process is as follows:

**Data Preparation:** The data preparation process involved two key stages: preprocessing and constructing. The following Figure 2 is the flowchart for this section:

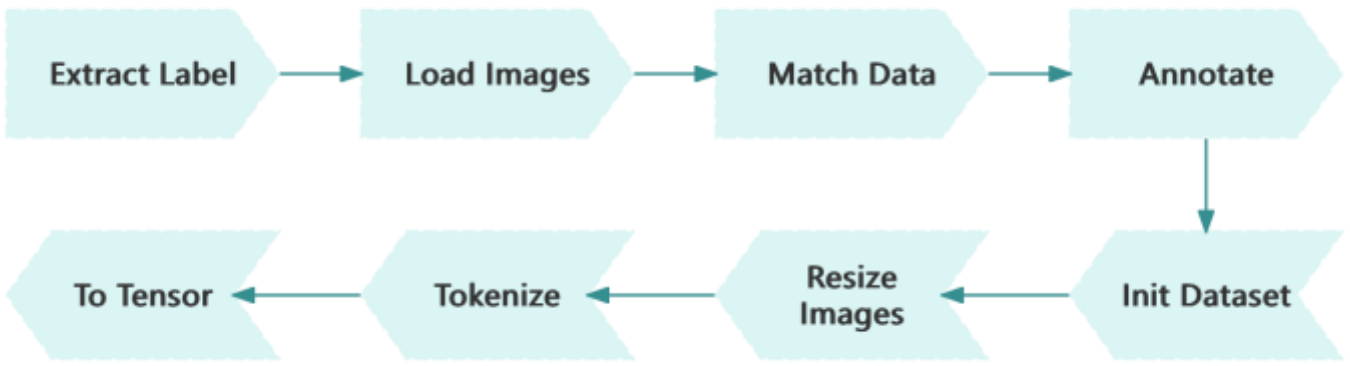

Figure 2 Data Preparation

During the preprocessing stage, the correct labels for the original images were first extracted from the CSV file. These labels provided the necessary supervision for the model during retraining. The adversarial samples were then retrieved from the designated image folder, where each image was systematically organized and labeled. This step ensured that the images were matched accurately with their corresponding labels, creating a structured and traceable dataset. An annotation file was generated to map each image to its corresponding label, establishing the groundwork for the custom dataset construction.

Following preprocessing, a custom dataset was created to optimize data handling during training. This dataset was initialized with the organized images, corresponding annotations, and a processing component specifically designed for the CLIP model. The dataset class was developed to efficiently retrieve, process, and prepare each sample for input into the model. Images were resized to a consistent dimension, normalized, and paired with their tokenized text labels, transforming them into tensors suitable for model input. This comprehensive data preparation pipeline ensured that the adversarial samples were correctly formatted and ready for effective use in the retraining process.

**Model Retraining:** The CLIP model is retrained using the expanded training set. A standard cross-entropy loss function is used, and learning rates and other training parameters are adjusted based on the model’s performance on the validation set. During training, the model’s performance on both clean and adversarial samples is closely monitored. After training, the fine-tuned model and its configurations are saved, ensuring easy future deployment and use.

**Performance Evaluation:** The model's performance was evaluated on a validation set consisting of adversarial samples paired with textual labels. The evaluation metrics included accuracy, precision, recall, and F1 score, providing a detailed assessment of the model's robustness. The results were visualized through bar charts, comparing the baseline and fine-tuned models, and a confusion matrix was generated to offer a comprehensive view of the model's performance across different classes.

## 3.2. Classifier training

The training process began with loading and preprocessing the CIFAR-10 test set using the CLIP model’s preprocessing pipeline, which included resizing the images to 224x224 pixels and normalizing them. Next, we implemented three adversarial attacks (DeepFool, FGSM, and AutoAttack) and extracted features from both the original and adversarially perturbed images. To ensure consistency in data input, these features were normalized and combined into a comprehensive feature set, encompassing adversarial samples generated by all attack methods.

After feature extraction and integration, we trained multiple models, including XGBoost, LightGBM, AdaBoost, and a deep-learning neural network model. The AdaBoost, XGBoost and LightGBM models were trained on the extracted features with an 80/20 split between training and validation sets, and hyperparameters such as learning rate and tree depth were fine-tuned using cross-validation. The neural network model was trained using stochastic gradient descent, with a focus on optimizing the model's performance in classifying adversarial examples. Finally, the trained models were evaluated on the test set using metrics such as accuracy, F1-score, precision, recall, and confusion matrix. These evaluations were computed using the sklearn library, providing a comprehensive analysis of each model's robustness against adversarial attacks.

# 4. EXPERIMENTS

## 4.1. Datasets

### 4.1.1. Retrain Clip

The dataset was divided into two main subsets: a training set and a validation set. The training set comprised primarily of adversarial examples, which were used to fine-tune the CLIP model. This subset was essential for teaching the model to recognize and correctly classify images even when they had been deliberately altered to cause misclassification. The validation set contained a mix of clean and adversarial examples, allowing for a comprehensive assessment of the model's performance post-training. The validation set was critical for evaluating the model's generalization capabilities, ensuring that it could handle both perturbed and unperturbed inputs effectively.

Below are examples from the training and validation datasets( Figure 3):

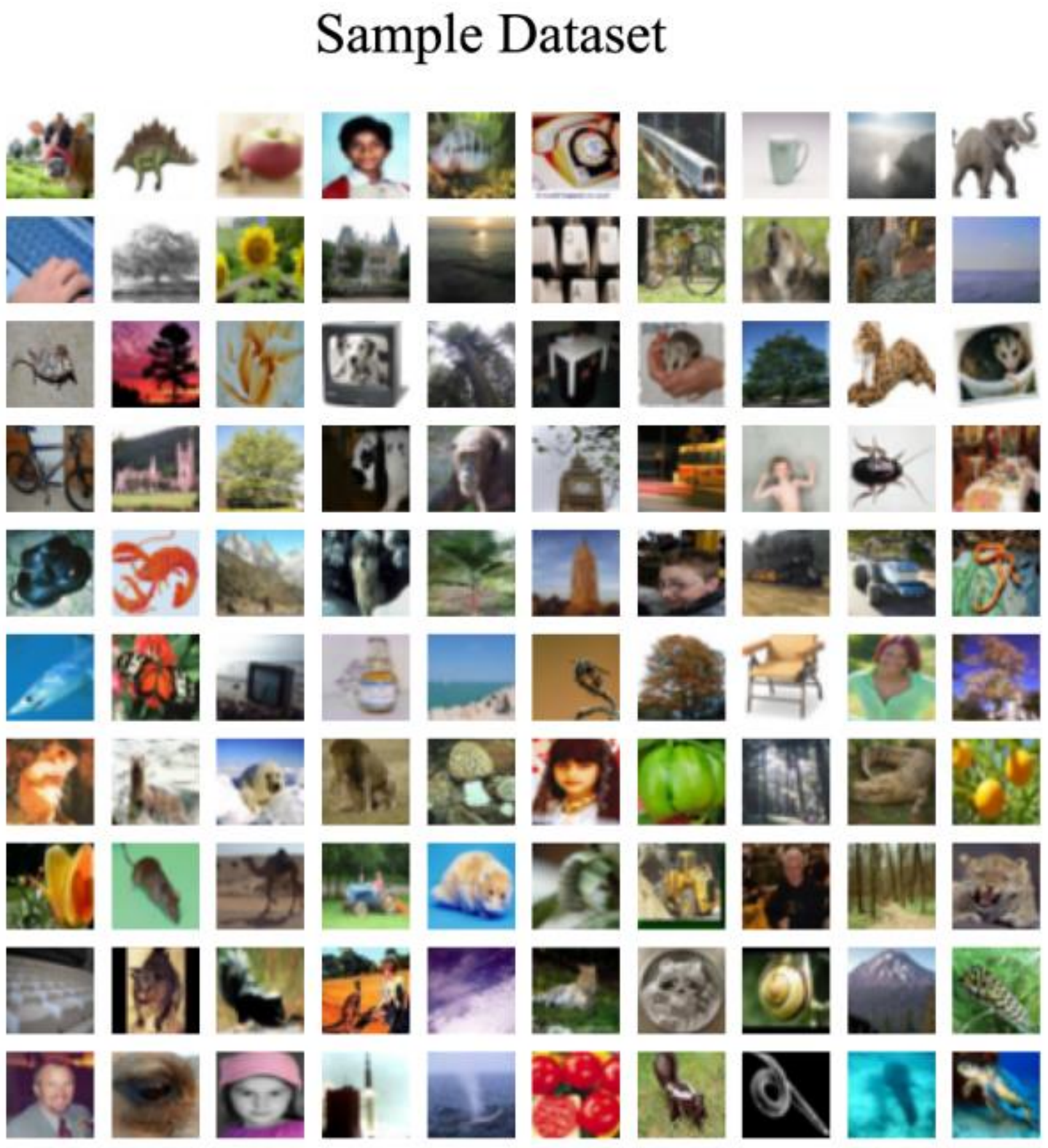

Figure 3 Sample Dataset for Training

#### 4.1.2. Classifier Datasets

For this study, we utilized the CIFAR-10 test set, consisting of 10,000 images across 10 classes. The images were preprocessed using the CLIP model's standard pipeline, which involved resizing and normalization to ensure compatibility with the model's input requirements.

In addition to the original images, we generated adversarial examples using Attack models such as DeepFool, FGSM, and AutoAttack. The CLIP model was then used to extract features from both the original and adversarially perturbed images. These features were combined into a comprehensive dataset, along with the true labels, to evaluate the robustness and accuracy of various machine learning models under adversarial conditions.

### 4.2. Experimental Setup

#### 4.2.1. Experimental Setup for Clip

The experiments were conducted on a computing environment natively equipped with a GPU to accelerate the training and evaluation processes. The fine-tuning and evaluation scripts were implemented in Python, using machine learning libraries such as PyTorch and the transformers library by Hugging Face.

### 4.3. Baselines

#### 4.3.1. Clip Model

In this experiment, the original CLIP model was employed as the baseline to evaluate the effectiveness of adversarial training. The vit-base-patch32 model, developed by OpenAI, is a vision-language model that integrates the Transformer architecture with visual feature extraction. It has been pretrained on a large-scale image-text paired dataset, enabling it to demonstrate robust performance in tasks involving both vision and language. The baseline model, having only undergone pretraining on a large-scale dataset without any exposure to adversarial examples, is expected to perform well on standard image classification tasks. However, due to the lack of specialized training against adversarial samples, it exhibits significant vulnerability when faced with carefully crafted adversarial attacks. The selection of vit-base-patch32 as the baseline model was driven by its widespread usage and recognized performance, making it an ideal reference point for assessing the improvements brought by adversarial training.

#### 4.3.2. Classifier Model

In our experiments, we employed four baseline machine learning models: AdaBoost, XGBoost, LightGBM, and Neural Networks. AdaBoost combines multiple weak classifiers into a strong one through iterative weight adjustments and weighted voting. XGBoost builds decision trees sequentially to minimize classification errors, leveraging cumulative predictions and learning rates. LightGBM grows trees leaf-wise, splitting the leaf with the highest information gain to optimize performance. Neural Networks, with their multi-layer architecture and non-linear transformations, use activation functions like ReLU to capture complex patterns in data, making them highly effective for tasks requiring deep learning.

## 5. EVALUATION

### 5.1. For Clip Retrain

The performance of both the baseline and fine-tuned models was rigorously evaluated using a set of standard metrics: accuracy, precision, recall, and F1 score. These metrics were selected for their ability to provide a comprehensive assessment of the model’s classification capabilities across various dimensions.

Accuracy was calculated as the proportion of correct predictions out of the total predictions, providing an overall measure of the model’s performance. Precision, defined as the ratio of true positive predictions to the total positive predictions, reflects the model’s ability to minimize false positives, while recall, the ratio of true positives to actual positives, indicates the model’s sensitivity in detecting true positives. The F1 score, as the harmonic mean of precision and recall, offers a balanced metric, particularly useful in cases where there is an imbalance between these two measures.

In addition to these metrics, a confusion matrix was generated to visually represent the distribution of correct and incorrect predictions across different classes. The confusion matrix offered detailed insights into the specific categories where the model performed well or struggled, allowing for a more granular analysis of its strengths and weaknesses.

The performance comparison between the baseline CLIP model and the fine-tuned model revealed significant improvements as a result of the adversarial training. The baseline model, which had only undergone pretraining on a standard large-scale dataset, exhibited a relatively low accuracy of 4% on the adversarial validation set. This low performance highlighted the baseline model’s vulnerability to adversarial attacks, as it was not specifically trained to handle such perturbations.

In contrast, the fine-tuned model achieved an accuracy of 43.5% on the same validation set, representing a substantial increase over the baseline. This improvement demonstrated the effectiveness of adversarial training in enhancing the model’s robustness. The precision, recall, and F1 scores also saw corresponding improvements, indicating that the model not only became better at correctly classifying adversarial examples but also maintained a more balanced performance across all evaluated metrics.

Figure 4 shows the Baseline Model Evaluation Metrics:

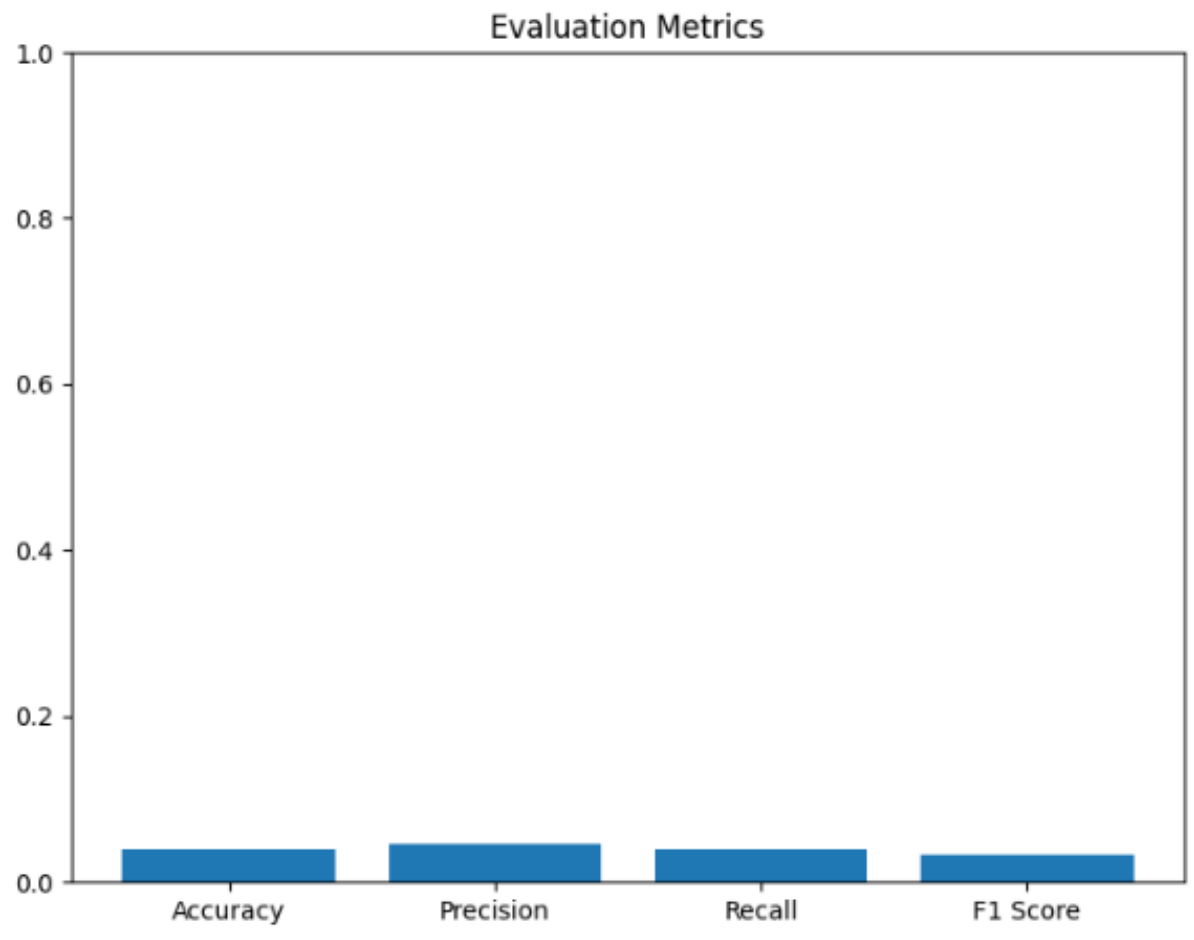

Figure 4 Baseline Evaluation Metrics

Figure 5 shows the Fine-Tuned Model Evaluation Metrics:

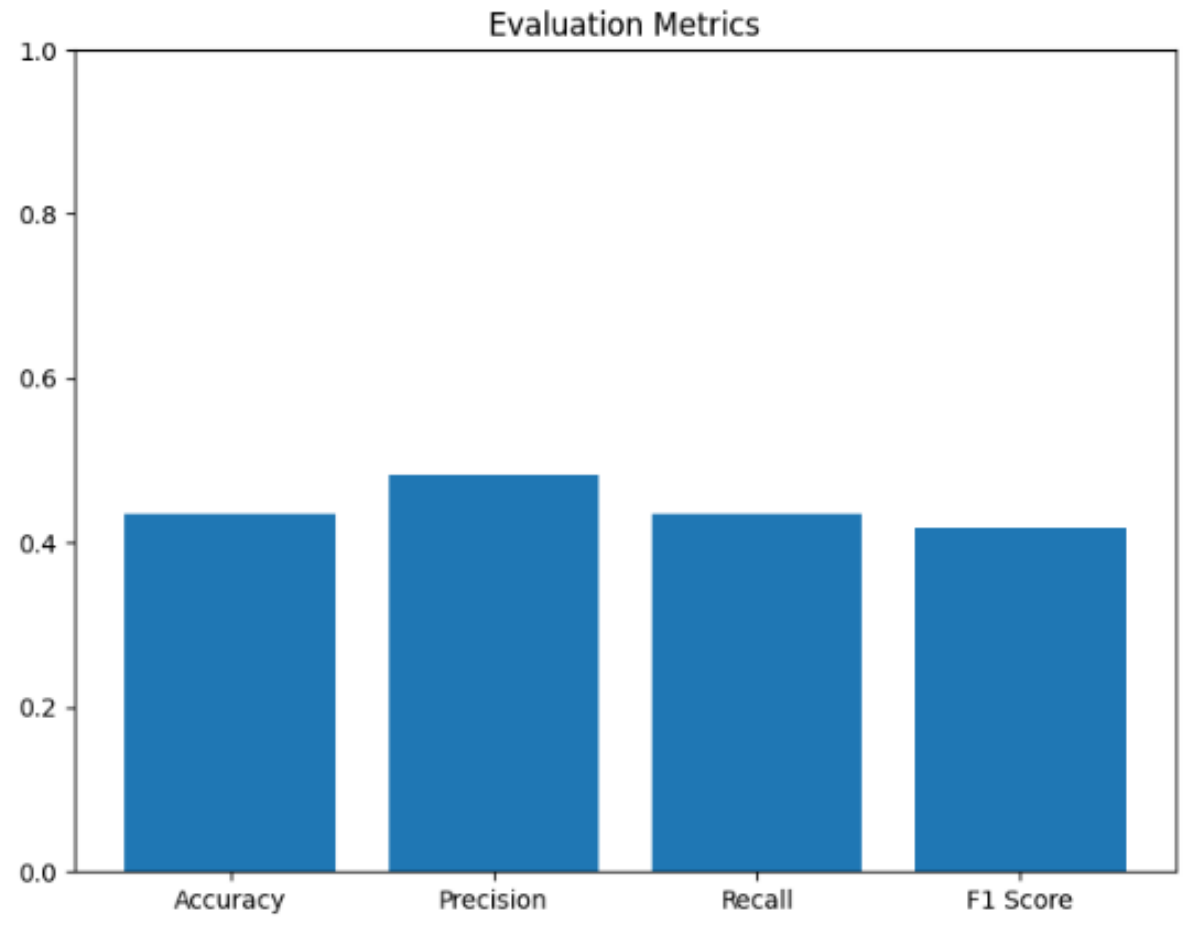

Figure 5 Fine-Tuned Evaluation Metrics

The confusion matrix further illustrated the differences in performance between the two models. For the baseline model, the confusion matrix showed a high degree of misclassification across almost all classes, reflecting the model's difficulty in correctly identifying adversarial inputs. In contrast, the confusion matrix for the fine-tuned model showed a much lower incidence of misclassification, with several classes showing significant improvement in correct predictions.

Figure 6 shows the Baseline Model Confusion Matrix:

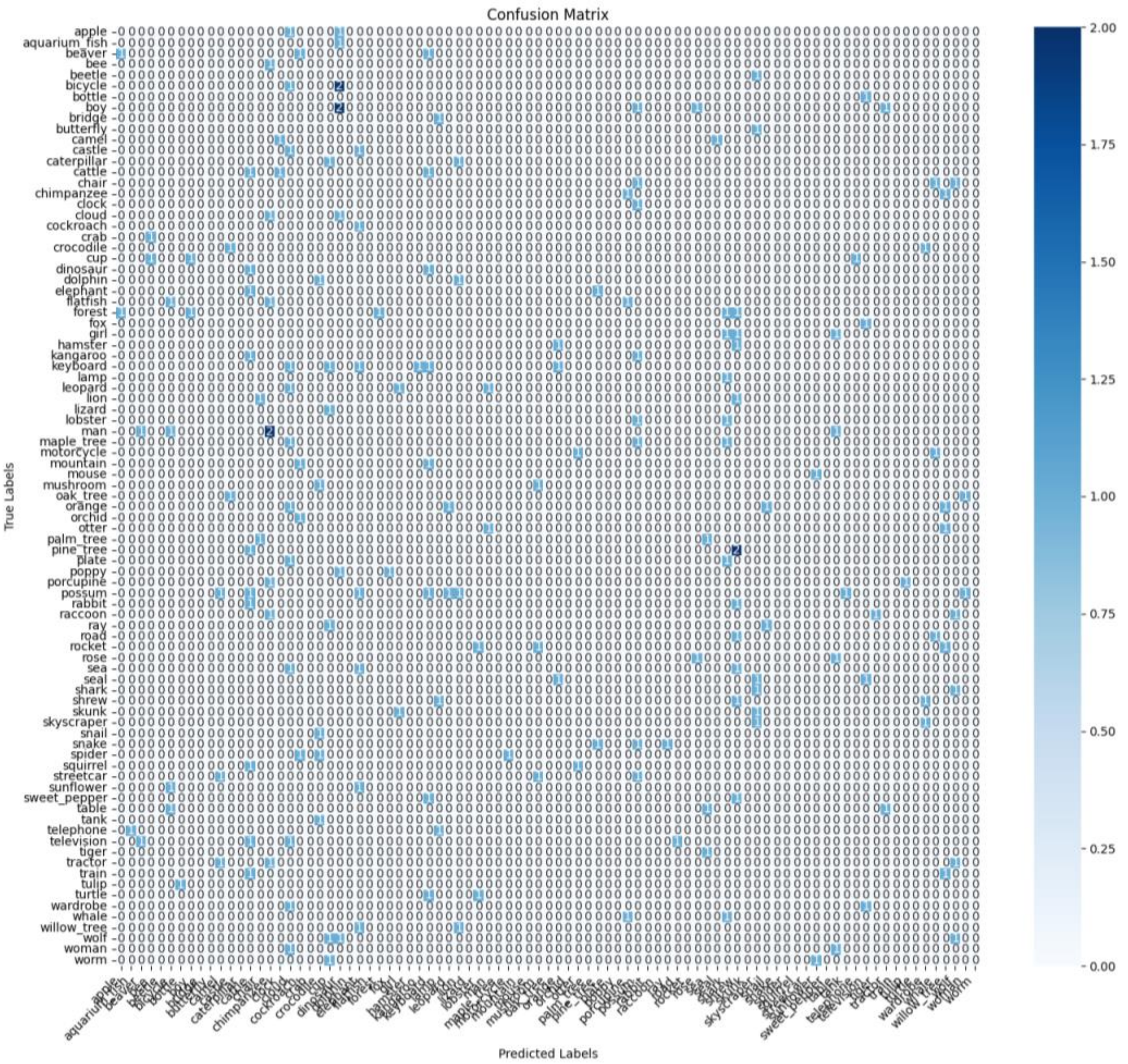

Figure 6 Baseline Confusion Matrix

Figure 7 shows the Fine-tuned Model Confusion Matrix:

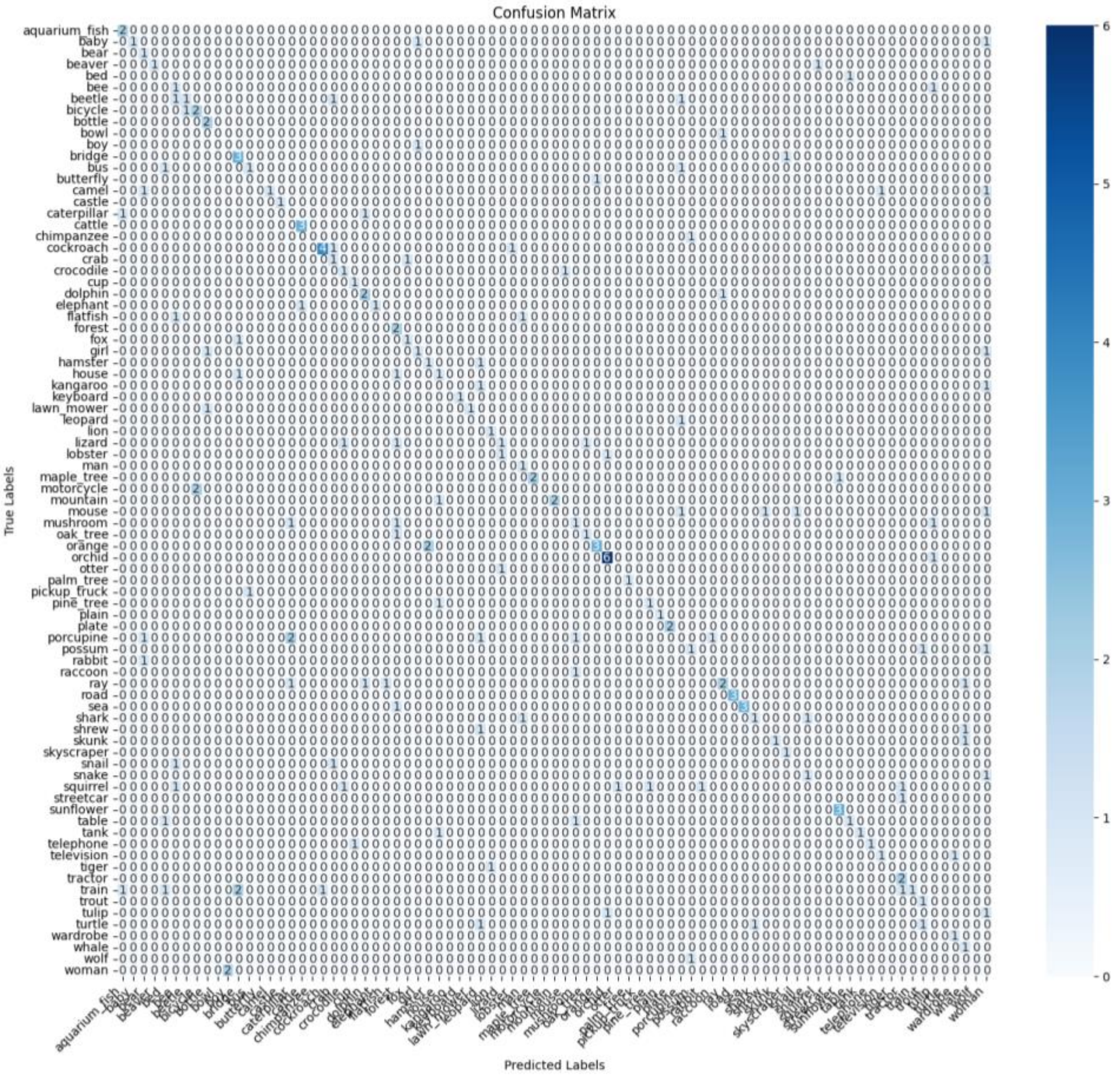

Figure 7 Fine-tuned Confusion Matrix

These results underscore the value of adversarial training as a method for improving the robustness of deep learning models, particularly in scenarios where security and accuracy are critical. The substantial increase in performance metrics from the baseline to the fine-tuned model suggests that the approach taken in this study effectively mitigates the risks posed by adversarial attacks, making the model more reliable for deployment in real-world applications where adversarial conditions might be present.

## 5.2. For Classifier Performance

### 5.2.1. Comparison

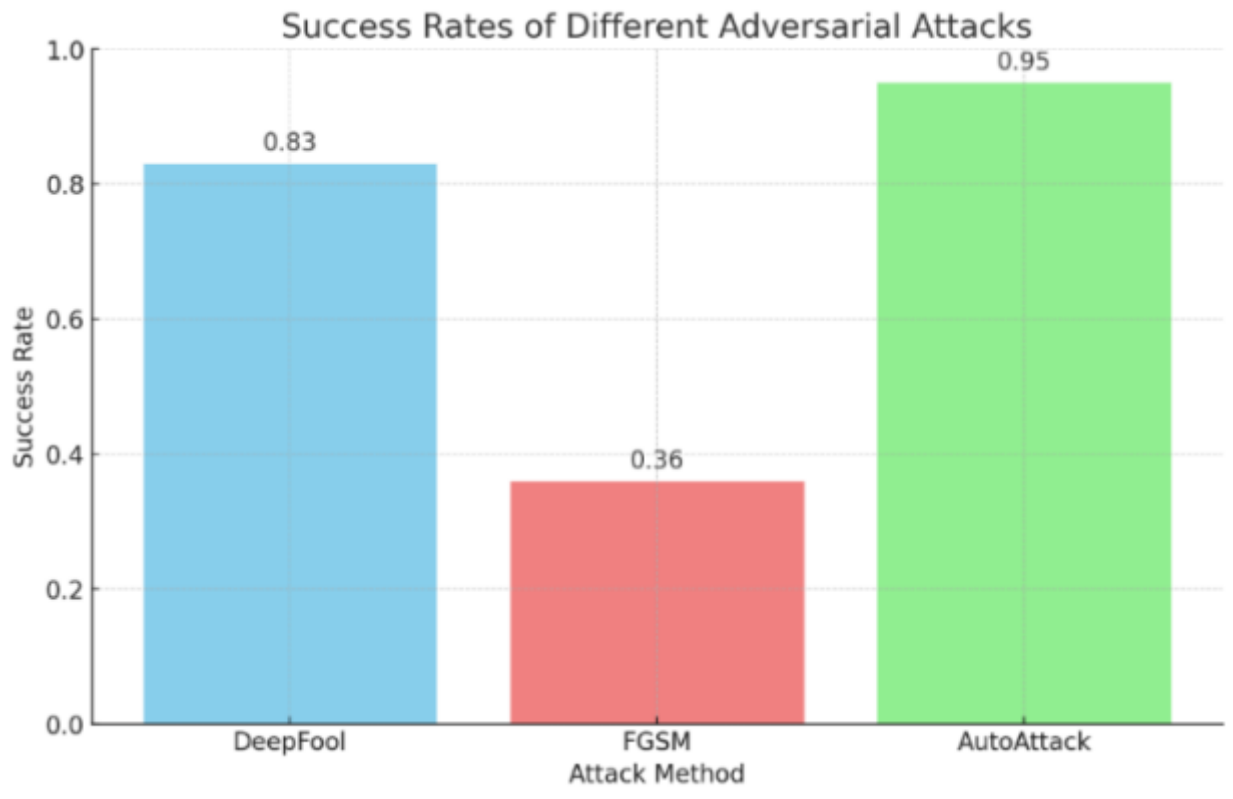

Figure 8 Adversarial Attack Success Rate

Here is a bar chart that illustrates the success rates of three different adversarial attack methods 8 (Figure 8). On the x-axis, we have the attack methods (DeepFool, FGSM, and AutoAttack), and on the y-axis, their corresponding success rates.

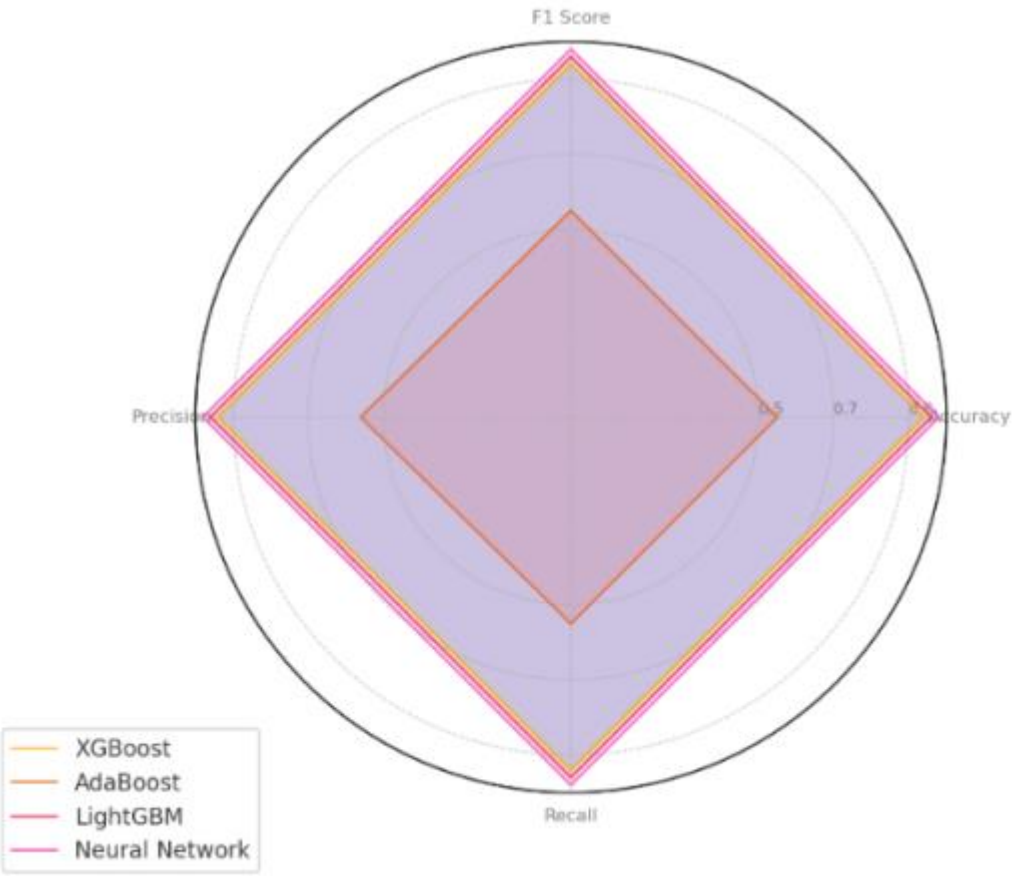

Figure 9 Accuracy of Four Models

The radar Figure 9 above compares the performance of four machine learning models (XGBoost, AdaBoost, LightGBM, and a Neural Network) across four metrics: Accuracy, F1 Score, Precision, and Recall.

XGBoost: Consistently performs with a score of 0.94 across all metrics, indicating balanced and strong performance.

AdaBoost: Shows lower and less consistent scores around 0.55 to 0.56, reflecting weaker performance compared to the other models.

LightGBM: Slightly outperforms XGBoost with scores of 0.96 across all metrics, highlighting its efficiency in handling the tasks.

Neural Network: Achieves the highest scores of 0.98 across all metrics, indicating the best performance among the models tested.

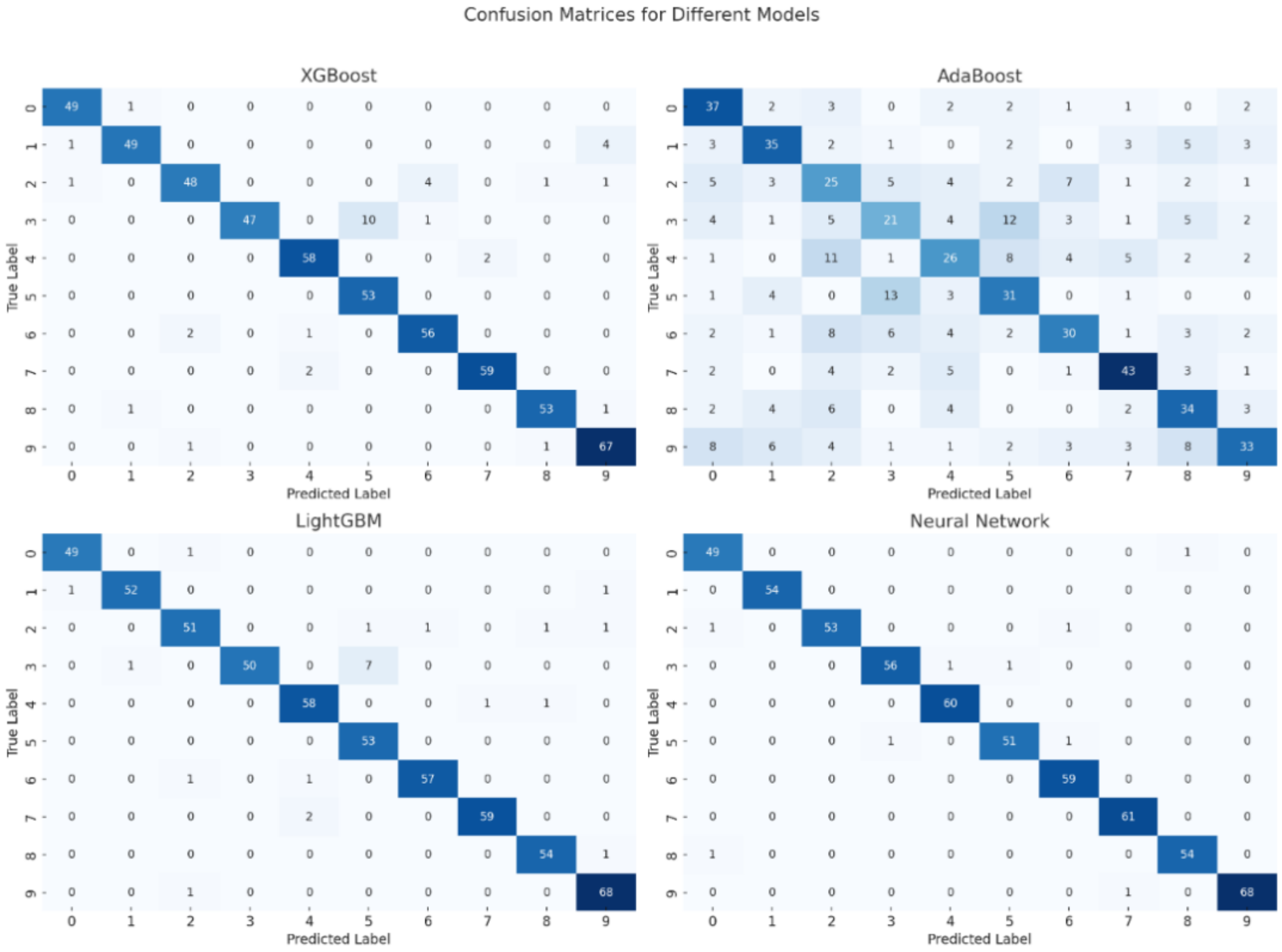

Figure 10 Confusion Matrices For Four Models

Figure 10 are the confusion matrices for the four models—XGBoost, AdaBoost, LightGBM, and Neural Network—presented as heatmaps. The intensity of the blue color represents the number of instances for each predicted versus true label combination. This visual representation allows for a clear comparison of each model's performance across different classes, making it easier to identify where each model excels or struggles in terms of classification accuracy.

The results show that LightGBM and the Neural Network outperformed XGBoost across various metrics, while AdaBoost significantly underperformed compared to the other models.

AdaBoost vs. XGBoost:

AdaBoost struggles due to its simplicity and sensitivity to noise and outliers, which are exacerbated by adversarial perturbations. Its weaker model complexity leads to poorer generalization and lower performance in adversarial settings.

XGBoost performs better by building more complex models with regularization to mitigate overfitting and handling noise more robustly.

LightGBM and Neural Network vs. XGBoost:

LightGBM benefits from its optimized leaf-wise tree growth and efficient handling of high-dimensional data, leading to better accuracy and robustness.

Neural Network excels in capturing the complex patterns in images, particularly in adversarial scenarios, due to its deep architecture and non-linear representation capabilities.

### 5.2.2. Classifier Sensitivity Analysis

In the sensitivity analysis, it was observed that the four models responded differently to changes in hyperparameters such as learning rate and tree depth. AdaBoost demonstrated extreme sensitivity to these changes, with performance significantly degrading even with minor adjustments. This sensitivity makes AdaBoost more prone to performance deterioration in adversarial environments. In contrast, XGBoost, LightGBM, and the Neural Network exhibited higher robustness against hyperparameter variations, with LightGBM and the Neural Network particularly maintaining high classification accuracy and stability across different settings.

### 5.2.3. Classifier In-depth Analysis

The in-depth analysis further examined the misclassification patterns of each model in adversarial settings. AdaBoost, due to its focus on hard-to-classify samples—which often includes adversarial examples—was more susceptible to misclassification under these conditions. While XGBoost performed reasonably well, it was outperformed by LightGBM and the Neural Network in handling complex data structures and adversarial noise. LightGBM excelled due to its optimized leaf-wise growth strategy, effectively managing complex data and adversarial perturbations. The Neural Network, with its deep architecture and non-linear representation capabilities, demonstrated superior performance in capturing intricate patterns in images, especially in adversarial scenarios. Overall, LightGBM and the Neural Network were the most effective in dealing with adversarial samples.

### 5.2.4. Prediction Accuracy and Misclassification Analysis in XGBoost Model

In this section, we provide a detailed visualization and analysis of the model's predictions on new data (images 3000 to 3100), focusing on prediction probabilities, success rate, and instances of misclassification. The model demonstrated a strong overall performance with a success rate of 85.26%, accurately predicting the majority of the image labels. As Figure 11 shows, the prediction probabilities revealed that the model maintained high confidence in its correct classifications, exemplified by a 99.54% confidence for Image 3000, which was correctly classified as class 5.

However, the analysis also identified several instances of misclassification, predominantly occurring between classes with overlapping visual features, such as classes 3 and 5 or classes 6 and 2. These errors suggest that the model struggles to differentiate between categories with similar characteristics. Additionally, lower confidence in some incorrect predictions, as seen in the misclassification of Image 3015 (true label 6, predicted as 2), reflects a degree of uncertainty in the model's decision-making process.

Furthermore, the model encountered challenges with images containing complex features or potential adversarial perturbations. This highlights areas where further improvements are necessary to enhance the model's robustness and accuracy in more challenging scenarios, ensuring better generalization and performance across a broader range of inputs.

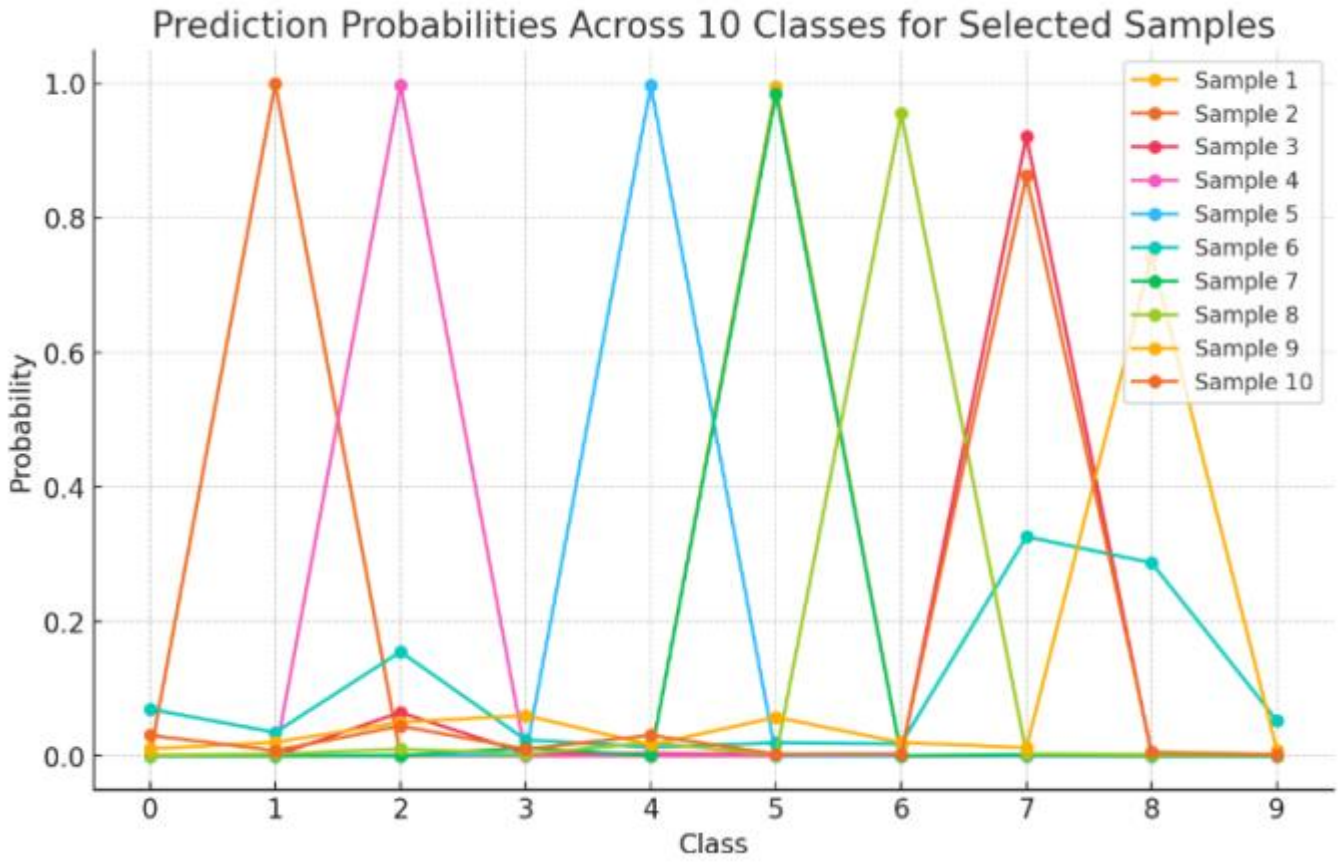

Figure 11 Prediction Probabilities

# 6. FUTURE WORKS

While our findings contribute significantly to the field of adversarial robustness, there are several avenues for future research:

**Exploration of Other Adversarial Attack Strategies:** Future studies could explore the application of additional adversarial attack strategies to further test and refine the robustness of VLMs. Incorporating more diverse adversarial examples could help in developing models that are even more resistant to various types of perturbations.

**Dynamic Adversarial Training:** Implementing dynamic adversarial training that adjusts the type and strength of adversarial examples during the training process could provide more robust defenses. By adapting in real-time to different adversarial strategies, models could learn to generalize better to unforeseen attacks.

**Integration of Advanced Machine Learning Models:** Further exploration into advanced machine learning models beyond XGBoost and LightGBM, such as CatBoost or more sophisticated neural network architectures, could enhance the robustness of VLMs. Research could focus on optimizing these models specifically for adversarial training to leverage their unique strengths fully.

# 7. CONCLUSION

In this study, we enhanced the robustness of the CLIP model against adversarial attacks. The experimental results demonstrate that the fine-tuned CLIP model significantly improved its ability to handle adversarial inputs, achieving an accuracy of 43.5% on adversarially perturbed images, compared to only 4% for the baseline model. This significant improvement, along with enhancements in precision, recall, and F1 score, highlights the effectiveness of our adversarial training framework.

We further analyzed the success rates of different attack methods and the performance of various models across different metrics. Specifically, the success rates of the three attack methods were as follows: DeepFool at 83%, FGSM at 36%, and AutoAttack at 95%. These results show that AutoAttack had the highest success rate, while FGSM was relatively weaker. These comparisons help us better understand the effectiveness of these attack methods and their threats to model robustness.

In terms of model performance comparison, LightGBM and Neural Networks performed exceptionally well across all metrics, achieving accuracies of 96% and 98%, respectively, significantly outperforming XGBoost and AdaBoost. It is worth noting that LightGBM, with its optimized leaf-wise growth strategy, not only achieved 96% accuracy but also excelled in handling high-dimensional data and countering adversarial perturbations. Neural Networks, on the other hand, were particularly outstanding in these challenging classification tasks, with F1 scores, precision, and recall all reaching 98%. In contrast, XGBoost achieved an accuracy of 94%, while AdaBoost, due to its sensitivity to noise and adversarial perturbations, performed relatively poorly across all metrics, with an accuracy of only 55%.

It is also noteworthy that the XGBoost model achieved a success rate of 85.26% when predicting new data (images 3000 to 3100), indicating strong overall performance on these data points. However, further analysis revealed that the model still faced challenges when dealing with images with complex features or potential adversarial perturbations, particularly in distinguishing between similar classes. This finding underscores the need for further improvements in model robustness and generalization capabilities in more challenging scenarios. In summary, the proposed method not only significantly improved the accuracy and robustness of the model under adversarial perturbations but also demonstrated superior performance in practical applications, providing a practical solution for enhancing model reliability.

## ACKNOWLEDGEMENT

Yijun Li, Qianhe Shen, Yuhan Liang and Yumeng Niu contributed equally to this work and should be considered co-first authors.

## REFERENCE

[1] Goodfellow, I. J., Shlens, J., & Szegedy, C. (2014). Explaining and Harnessing Adversarial Examples. arXiv preprint arXiv:1412.6572.

[2] Madry, A., Makelov, A., Schmidt, L., Tsipras, D., & Vladu, A. (2018). Towards deep learning models resistant to adversarial attacks. In International Conference on Learning Representations.

[3] Athalye, A., Carlini, N., & Wagner, D. (2018). Obfuscated gradients give a false sense of security: Circumventing defenses to adversarial examples. In *Proceedings of the 35th International Conference on Machine Learning* (Vol. 80, pp. 274-283).

[4] Chen, T., & Guestrin, C. (2016). XGBoost: A scalable tree boosting system. In Proceedings of the 22nd ACM SIGKDD International Conference on Knowledge Discovery and Data Mining.

[5] Ke, G., Meng, Q., Finley, T., Wang, T., Chen, W., Ma, W., ... & Liu, T.-Y. (2017). LightGBM: A highly efficient gradient boosting decision tree. In Advances in Neural Information Processing Systems (pp. 3146-3154).

[6] Moosavi-Dezfooli, S. M., Fawzi, A., & Frossard, P. (2016). Deepfool: a simple and accurate method to fool deep neural networks. In Proceedings of the IEEE conference on computer vision and pattern recognition (pp. 2574-2582).

[7] Croce, F., & Hein, M. (2020). Reliable evaluation of adversarial robustness with an ensemble of diverse parameter-free attacks. Proceedings of the 37th International Conference on Machine Learning (ICML), PMLR, 1576-1586.

[8] Radford, A., Kim, J. W., Hallacy, C., Ramesh, A., Goh, G., Agarwal, S., ... & Sutskever, I. (2021). Learning Transferable Visual Models From Natural Language Supervision. In Proceedings of the 38th International Conference on Machine Learning (pp. 8748-8763).